\documentclass{article}

    \usepackage[preprint, nonatbib]{neurips_2020}

\usepackage[utf8]{inputenc} 
\usepackage[T1]{fontenc}    
\usepackage{hyperref}       
\usepackage{url}            
\usepackage{booktabs}       
\usepackage{amsfonts}       
\usepackage{nicefrac}       
\usepackage{microtype}      

\usepackage{graphicx}
\usepackage{amsmath}
\usepackage[section]{placeins}

\newcommand{\ra}[1]{\renewcommand{\arraystretch}{#1}}

\DeclareMathOperator*{\argmax}{arg\,max}

\title{Generalized SHAP: Generating multiple types of explanations in machine learning}

\author{%
  Dillon Bowen\thanks{Dillon Bowen is credited with the original concept, code, data analysis, and initial paper draft. Lyle Ungar is credited with contributions to the draft and mathematical notation. Documentation can be found at https://dsbowen.github.io/gshap/.} \\
  Operations, Information, and Decisions Department\\
  The Wharton School of Business, University of Pennsylvania\\
  Philadelphia, PA 19104 \\
  \texttt{dsbowen@wharton.upenn.edu} \\
  \And
  Lyle Ungar \\
  Department of Computer and Information Science, University of Pennsylvania \\
  Philadelphia, PA 19104 \\
  \texttt{ungar@cis.upenn.edu} \\
}

\begin{document}

\maketitle

\begin{abstract}
    Many important questions about a model cannot be answered just by explaining how much each feature contributes to its output. To answer a broader set of questions, we generalize a popular, mathematically well-grounded explanation technique, Shapley Additive Explanations (SHAP). Our new method - Generalized Shapley Additive Explanations (G-SHAP) - produces many additional types of explanations, including: 1) \textit{General classification explanations}; Why is this sample more likely to belong to one class rather than another? 2) \textit{Intergroup differences}; Why do our model's predictions differ between groups of observations? 3) \textit{Model failure}; Why does our model perform poorly on a given sample? We formally define these types of explanations and illustrate their practical use on real data.
\end{abstract}














\section{Introduction}

It is increasingly important to explain how machine learning models work \cite{lipton2018mythos}. For example, machine learning models are currently used to assess criminals' risk of recidivism \cite{compas}, assist in medical diagnosis \cite{davenport2019potential}, and recruit employees \cite{datta2015automated}. Researchers have proposed a variety of methods to measure which features most affect a model's output \cite{deeplift, shap, lime}. However, many important questions about a model cannot be answered just by measuring feature importance with respect to its output. Rather, we need to measure feature importance with respect to {\em some function} of the model's output. Considering the following examples:

\paragraph{Example 1: General classification.} Suppose we have a black-box model which diagnoses patients with COVID-19, the flu, or the common cold based on their symptoms. Existing explanatory methods can tell us why our model diagnosed a patient with COVID-19. However, medical practitioners, public health officials, and patients may also want to know, \textit{why was this patient diagnosed with COVID-19, rather than the flu?} and \textit{how do the symptoms which distinguish COVID-19 from the flu differ from those which distinguish COVID-19 from the common cold?}. These ‘general classification explanations’ are not provided by existing explainable AI methods.

\paragraph{Example 2: Intergroup differences.} Suppose we have a black-box model which predicts a criminal’s risk of recidivism to determine whether they are eligible for parole. Existing explanatory methods can tell us why our model predicted that a criminal has a high recidivism risk. However, lawyers and criminal justice policymakers may also want to know why our model predicts that Black criminals generally have a higher recidivism risk than White criminals. These ‘intergroup difference explanations’ are also not handled cleanly by existing feature importance methods.

\paragraph{Example 3: Model failure.} Suppose we have a black-box model which forecasts GDP growth based on macroeconomic variables. Existing explanatory methods can tell us why our model forecast 3\% GDP growth in a given year. However, fiscal policymakers, financial analysts, and central bankers may also want to know why our model failed to predict the 2008-2009 Great Recession. Such `model failure' explanations are also not cleanly provided by existing methods.

To answer these questions, we generalize a popular, mathematically well-grounded explanation technique, Shapley Additive Explanations (SHAP). Our new method - Generalized Shapley Additive Explanations (G-SHAP) - produces three new types of explanations which current techniques do not cleanly provide: general classification explanations, intergroup difference explanations, and model failure explanations (see Table \ref{tab:explanations}). We formally define these types of explanations and illustrate their practical use on real data.

\begin{table}
\centering
\caption{Types of explanations generated by G-SHAP}
\ra{2}
\begin{tabular}{@{}lp{10cm}@{}}
\toprule
Explanation type & General formulation \\
\midrule

{\bf General classification} \\
\hspace{1em}  Question
    & Let $f_c(x_i)$ be the probability that observation $x_i$ belongs to class $c$ according to model $f$. Why are the observations in set $X$ more likely to belong to a positive class $c\in C_+$ than a negative class $c\in C_-$?
    \\
\hspace{1em} $g(f,X,\Omega)$
    & $\frac{\prod_i \sum_{c\in C_+} f_c(x_i)}
    {\prod_i \sum_{c\in C_+} f_c(x_i) + \prod_i \sum_{c\in C_-} f_c(x_i)}$
    \\
\hspace{1em} $\Omega$
    & $\{C_+,C_-\}$
    \\
\hspace{1em} Explanation
    & Let $Pr_X$ [$Pr_Z$] be the probability that every observation in set $X$ [the background set $Z$] belongs to a positive class rather than a negative class. $\phi_g^j$ is the number of percentage points of this difference $Pr_X-Pr_Z$ explained by feature $j$.
    \\

{\bf Intergroup differences} \\
\hspace{1em} Question
    & Why is there a difference $d$ between the model output for $X_0\subseteq X$ and the model output for $X_1\subseteq X$?
    \\
\hspace{1em} $g(f,X,\Omega)$
    & $d(E_{x_i\in X_0}[f(x_i)],E_{x_i\in X_1}[f(x_i)])$
    \\
\hspace{1em} $\Omega$
    & $\{X_0,X_1\}$
    \\
\hspace{1em} Explanation
    & $\phi_g^j$ is the amount of the intergroup difference explained by feature $j$.
    \\

{\bf Model failure} \\
\hspace{1em} Question
    & Why did our model perform poorly on a set of observations $X$ with labels $y$, as measured by loss function $L$?
    \\
\hspace{1em} $g(f,X,\Omega)$
    & $L(f,X,y)$
    \\
\hspace{1em} $\Omega$
    & $\{y\}$
    \\
\hspace{1em} Explanation
    & Let $L(f,X,y)$ [$L(f,Z,y)$] be the loss over a set of observations $X$ [the background set $Z$]. $\phi_g^j$ is the amount of this difference $L(f,X,y)-L(f,Z,y)$ explained by feature $j$.
    \\
\bottomrule
\end{tabular}
\end{table}
\label{tab:explanations}


\section{Generalized Shapley Additive Explanations}

Researchers have proposed several methods to measure which features most affect a model's output \cite{deeplift,shap,lime}. Shapley Additive Explanations (SHAP), derived from Shapley values in game theory \cite{shapley1953value,roth1988introduction}, is a popular and mathematically well-grounded example of a feature importance measure \cite{cohen2005feature,shap}. According to SHAP, the importance of feature $j$ for the output of model $f$, $\phi^j(f)$, is a weighted sum of the feature's contribution to the model's output $f(x_i)$ over all possible feature combinations \cite{molnar2019}:

\begin{equation}
    \phi^j(f) = \sum_{S\subseteq \{x^1,...,x^p\}\setminus\{x^j\}}
        \frac{|S|!(p-|S|-1)!}{p!}\big(f(S\cup \{x^j\})-f(S)\big)
\end{equation}

Where $x^j$ is feature $j$, $S$ is a subset of features, and $p$ is the number of features in the model.

In practice, $f(S)$ is estimated by substituting in values for the remaining features, $\{x^1,…,x^p\}\setminus S$, from a randomly selected observation in a background dataset $z\in Z$. Suppose we compute the model output for an observation $f(x_i)$ and a background observation $f(z)$. Each SHAP value $\phi^j$ is the amount of this difference $f(x_i)-f(z)$ due to feature $j$.

Shapley feature importance measures have many practical applications. These applications include using SHAP to identify patient risk factors in tree-based medical diagnostic models \cite{lundberg2019explainable} and determine which features of satellite images are most important in generating poverty maps \cite{ayush2020generating}. Some researchers have modified Shapley values for specific use cases. For example, \cite{singal2019shapley} uses counterfactual-adjusted Shapley values for attributing sales to advertisements, and \cite{jha2019attribution} uses a Shapley variant to measure which features lead deep neural networks to be `confident' in their output.

Here we present a unifying framework for SHAP and its variants; Generalized Shapley Additive Explanations (G-SHAP). G-SHAP allows us to compute the feature importance of any function $g$ of a model's output. Define a G-SHAP value $\phi_g^j(f)$ as:

\begin{equation}
    \phi_g^j(f) = \sum_{S\subseteq \{x^1,...,x^p\}\setminus\{x^j\}}
        \frac{|S|!(p-|S|-1)!}{p!}\big(g(f,S\cup \{x^j\},\Omega)-g(f,S,\Omega)\big)
\end{equation}

Where $\Omega$ is a set of additional arguments.

G-SHAP values have a similar interpretation as SHAP values. Suppose we compute a function of a model’s output for a sample $g(f,X,\Omega)$ and a background dataset $g(f,Z,\Omega)$. Each G-SHAP value $\phi_g^j$ is the amount of this difference $g(f,X,\Omega)-g(f,Z,\Omega)$ due to feature $j$.

In the following sections, we demonstrate how to use G-SHAP, along with select generalized functions $g$, to produce three new types of explanations not provided by existing techniques: general classification explanations, intergroup difference explanations, and model failure explanations.

\section{General classification explanations}

\paragraph{Example.} In this example, we ask how the features which distinguish Versicolour from Setosa irises compare to those which distinguish Versicolour from Virginica irises. (This is analogous to asking how the symptoms which distinguish COVID-19 from the flu compare to those which distinguish COVID-19 from the common cold; however, a sufficient COVID-19 dataset is not publicly available.) We find that petal length is the most important feature. Additionally, petal length is relatively less important, and sepal length relatively more important, for distinguishing Versicolour from Setosa (as opposed to distinguishing Versicolour from Virginia).

\paragraph{General formulation.} More generally, let $f_c(x_i)$ be the probability that observation $x_i$ belongs to class $c\in C$ according to model $f$. Let $C_+\subset C$ be a set of `positive classes' and $C_-\subset C$ be a set of `negative classes'. For a single observation $x_i$, we can ask why our model predicts it comes from a positive class rather than a negative class. For a randomly selected set of observations $X$ from positive classes, we can ask which features best distinguish observations which belong to positive classes from observations which belong to negative classes. Here, the positive class is Versicolour, and the negative class is Setosa or Virginica.

\paragraph{Model.} We train Scikit-learn's support vector classifier with default settings on the Scikit-learn iris dataset ($n=150$), using a 75\%-25\% train-test split \cite{scikit-learn}. The purpose of this dataset is to predict an iris's species (Versicolour, Setosa, or Virginica) based on its petal length, petal width, sepal length, and sepal width. Our classifier achieves 97\% test accuracy.

\paragraph{G-SHAP values.} We compute the G-SHAP values, where $g$ measures the probability that every observation in set $X$ belongs to a positive class given that every observation belongs to a positive class or every observation belongs to a negative class.

\begin{equation}
g(f,X,\Omega) = \frac{\prod_i \sum_{c\in C_+} f_c(x_i)}{\prod_i \sum_{c\in C_+} f_c(x_i) + \prod_i \sum_{c\in C_-} f_c(x_i)}
\end{equation}

Where the additional arguments to $g$ are the sets of positive and negative classes; $\Omega=\{C_+,C_-\}$.

Note that traditional SHAP values are a specific instance of $g(f,X,\Omega)$ in which set $X$ contains a single observation $x_i$, the set of positive classes is the model's classification of $x_i$; $C_+=\{\argmax_{c\in C} f_c(x_i)\}$, and the negative classes are the complement of the positive classes $C_-=C\setminus C_+$. Also note that, while our generalized function $g(f,X,\Omega)$ uses only two subsets of classes, $C_+$ and $C_-$, it easily extends to accommodate any number of subsets.

In our example, $g$ measures the probability that every iris in this sample is a Versicolour, given that every iris in this sample is a Versicolour or every iris in this sample is a Setosa. Specifically, the set of positive classes $C_+$ is $\{\text{Versicolour}\}$, and the set of negative classes $C_-$ is $\{\text{Setosa}\}$.

\begin{equation}
g(f,X,\Omega)
= \frac{\prod_i f_{\text{Versicolour}}(x_i)}
{\prod_i f_{\text{Versicolour}}(x_i) + \prod_i f_{\text{Setosa}}(x_i)}
\end{equation}

Similarly, to discover which features distinguish Versicolour from Virginica, $C_-$ is $\{\text{Virginica}\}$. As before, $C_+=\{\text{Versicolour}\}$.

\begin{figure}
    \centering
    \includegraphics[scale=.7]{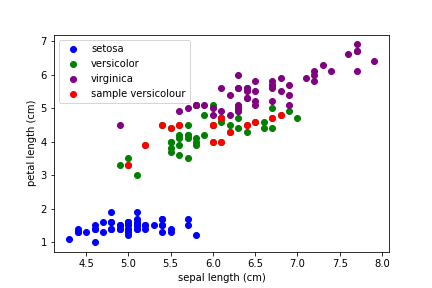}
    \caption{Plot of iris sepal length and petal length by species. Our sample Versicolour irises are highlighted in red.}
    \label{fig:iris_scatter}
\end{figure}

\paragraph{Interpretation.} To discover which features distinguish Versicolour from Setosa or Virginica, our set of observations $X$ is a random sample of test observations classified as Versicolour (Figure \ref{fig:iris_scatter}). Note that our analysis will not work if $X$ contains a mixture of iris species. Intuitively, we are asking what features lead our model to believe that $X$ contains only Versicolour irises. If $X$ contains a mixture of iris species, the G-SHAP values will be 0 for all features, indicating that our model does not believe $X$ contains only Versicolour irises.

For interpretation, we compare our Versicolour sample $X$ to a background dataset $Z$ (the shuffled training data) in terms of $g$. For the Versicolour-Setosa comparison, our support vector classifier correctly predicts a 100\% chance that our sample contains only Versicolour, as opposed to only Setosa (see Table \ref{tab:iris_comparison}). It also predicts a 65\% chance that our background dataset contains only Versicolour, as opposed to only Setosa. Each G-SHAP value $\phi_g^j$ is the number of percentage points of this difference (35 percentage points) explained by feature $j$. We interpret the G-SHAP values for the Versicolour-Virginica comparison similarly.

\begin{table}[]
    \centering
    \caption{Comparing Versicolour sample $X$ to background data $Z$ in terms of $g$}
    \ra{1.3}
    \begin{tabular}{@{}lrr@{}}
        \toprule
        & \multicolumn{2}{c}{Comparison} \\
        \cmidrule{2-3}
        & Versicolour-Setosa & Versicolour-Virginica \\
        \midrule
        $g(f,X,\Omega)$ & 1.00 & 1.00 \\
        $g(f,Z,\Omega)$ & 0.65 & 0.42 \\
        $g(f,X,\Omega)-g(f,Z,\Omega)$ & 0.35 & 0.58 \\
        \bottomrule
    \end{tabular}
    \label{tab:iris_comparison}
\end{table}

For general classification explanations more broadly, our model predicts that each observation in our sample belongs to a positive class, rather than a negative class, with probability $Pr_X$. It also predicts that each observation in the background dataset comes from a positive class, rather than a negative class, with probability $Pr_Z$. Each G-SHAP value $\phi_g^j$ is the number of percentage points of this difference $Pr_X-Pr_Z$ explained by feature $j$.

\begin{figure}
    \centering
    \includegraphics[scale=0.7]{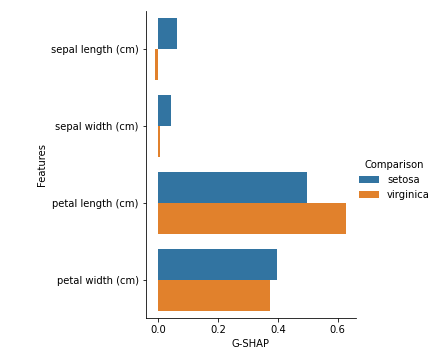}
    \caption{Normalized G-SHAP values for distinguishing Versicolour from Setosa (blue) and Versicolour from Virginica (orange). Note that petal length is relatively more important for the latter distinction, and sepal width relatively more important for the former.}
    \label{fig:iris_gshap}
\end{figure}

\paragraph{Results.} We plot our results in Figure \ref{fig:iris_gshap}, normalizing the G-SHAP values to sum to 1 for both comparisons (Versicolour-Setosa and Versicolour-Virginica). We find that petal length is the most important feature. Additionally, petal length is relatively less important, and sepal length relatively more important, for distinguishing Versicolour from Setosa (as opposed to distinguishing Versicolour from Virginia).

\section{Intergroup differences}

\paragraph{Related work.} Machine learning models are often used to study intergroup differences. For example, \cite{lundberg2019explainable} describes a method to identify groups which differ in feature importances with respect to a model's output. The method begins by computing feature importances with respect to a model's output using SHAP, then clusters observations in a feature importance embedding space. By contrast, we present a method to identify features which explain why a model's output differs between groups. Our method begins with groups of observations for which a model makes different predictions, then computes feature importances with respect to the difference.

\paragraph{Example.} In this example, we create a model to predict whether a criminal will recidivate within two years of release. We find that our model predicts higher recidivism rates for Black criminals than for White criminals, then use G-SHAP to identify which features are most responsible for this difference. We find that prior convictions, age, and race are the most important variables.

\paragraph{General formulation.} More generally, suppose we have two groups of observations, $X_0\subseteq X$ and $X_1\subseteq X$, and a measure $d$ of how different our model output is between groups. We can ask why our model makes different predictions for different groups. Here, groups $X_0$ and $X_1$ are White and Black criminals respectively, the model output is a classification which indicates that a criminal will recidivate within two years, and the difference measure is the relative difference in recidivism rates.

\begin{figure}
    \centering
    \includegraphics[scale=.7]{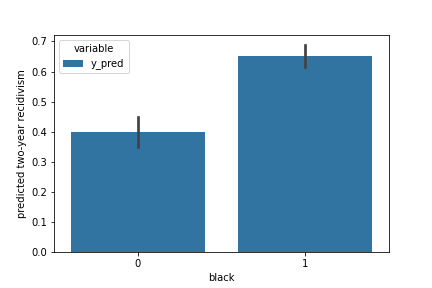}
    \caption{Difference in predicted recidivism rates between White and Black criminals.}
    \label{fig:recidivism_difference}
\end{figure}

\paragraph{Model.} We train Scikit-learn's support vector classifier \cite{scikit-learn} with default settings on the Correctional Offender Management Profiling for Alternative Sanctions dataset \cite{propublica}, selecting only Black and White criminals ($n_\text{Black}=3565,n_\text{White}=1904$), using a 75\%-25\% train-test split. The purpose of this dataset is to predict whether a criminal will recidivate within two years of release based on demographic information and criminal history. Our classifier achieves 68\% test accuracy. The predicted recidivism rate is 63\% higher for Black criminals (65\%) than for White criminals (40\%) (Figure \ref{fig:recidivism_difference}).

\begin{figure}
    \centering
    \includegraphics[scale=.7]{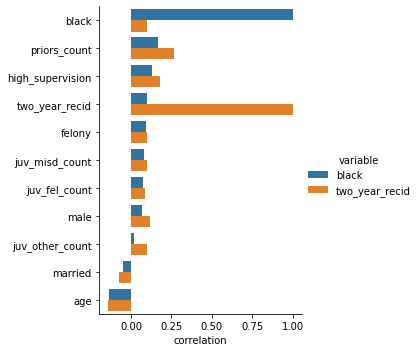}
    \caption{Plot of the correlations of all variables in our dataset with race (Black) and two-year recidivism.}
    \label{fig:recidivism_corr}
\end{figure}

Figure \ref{fig:recidivism_corr} plots the correlations of all variables in our dataset with race (Black) and two-year recidivism. Visually, it is unclear which variables lead our model to believe that Black criminals are at higher risk for recidivism.

\paragraph{G-SHAP values.} We compute the G-SHAP values, where $g$ measures the difference in model outputs between groups. In this case, $d$ is the relative difference between group means. Intuitively, this measures the relative difference in predicted recidivism rates between Black and White criminals. In the binary case, relative mean difference is known as \textit{disparate impacts} in the algorithm fairness literature \cite{feldman2015certifying}.

\begin{equation}
g(f,X,\Omega)
= d(E_{x\in X_0}[f(x)],E_{x\in X_1}[f(x)])
= \frac{n_\text{Black}^{-1}\sum_{x_i\in X_\text{Black}}\mathbf{1}(f(x_i)=\text{Recidivate})}
{n_\text{White}^{-1}\sum_{x_i\in X_\text{White}}\mathbf{1}(f(x_i)=\text{Recidivate})}
-1
\end{equation}

Where additional arguments to $g$ indicate group membership; $\Omega=\{X_0,X_1\}$.

Although we express the intergroup difference measure as a function of group means, our method easily extends to intergroup difference measures more generally; for example, a difference between group medians or a Wasserstein distance between sample distributions.

\paragraph{Interpretation.} For interpretation, we compare our test data $X$ to a background dataset $Z$ (the shuffled training data) in terms of $g$. Consistent with our previous result, the relative difference in predicted recidivism rates is 63\%. The relative difference for our shuffled background data is 0\% (because the background data are shuffled, the model has no way to statistically discriminate between groups). Each G-SHAP value $\phi_g^j$ is the number of percentage points of this difference (63 percentage points) explained by feature $j$.

For intergroup difference explanations more broadly, our model predicts the distance between groups is $d(E_{x\in X_0}[f(x)],E_{x\in X_1}[f(x)])$ for our sample, and 0 for the shuffled background data. Each G-SHAP value $\phi_g^j$ is the amount of this difference explained by feature $j$.

\begin{figure}
    \centering
    \includegraphics[scale=.7]{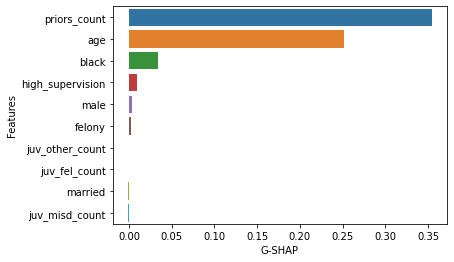}
    \caption{G-SHAP values for why our model predicts higher recidivism rates for Black criminals than for White criminals.}
    \label{fig:recidivism_gshap}
\end{figure}

\paragraph{Results.} We plot our results in Figure \ref{fig:recidivism_gshap}. This plot tells us which features explain the relative difference in predicted recidivism rates between Black and White criminals. We see that the most important features are number of prior convictions, age, and race.

\section{Model failure}

\paragraph{Related work.} Models which perform well during training often perform poorly in real-world applications. The ubiquity of this problem is reflected in recent efforts to make machine learning more adaptable and robust. These efforts include transfer learning \cite{pan2010transfer} and lifelong learning \cite{chen2018lifelong}. When, however, our models fail, we demand to know why. In the wake of the 2008-2009 financial crisis, for example, politicians, central bankers, and ordinary citizens demanded to know why economic models failed to predict the most serious economic downturn since the Great Depression.

\paragraph{Example.} In this example, we create a model to forecast GDP growth based on macroeconomic variables, then ask why it fails to predict the 2008-2009 financial crisis. We find that a high volume of real estate loans as a percent of GDP in 2007-2008 was detrimental to model performance.

\paragraph{General formulation.} More generally, we can ask why our model performed poorly on sample $X$, as measured by loss function $L$. Here, our sample is the years 2008-2009, and our loss function is $R^2$.

\begin{figure}
    \centering
    \includegraphics[scale=.7]{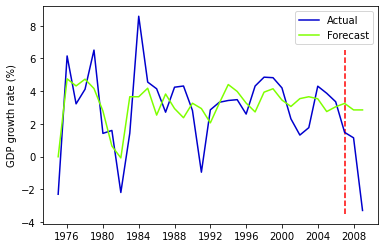}
    \caption{GDP growth forecasts from 1975 to 2009. Our model was trained on years 1975-2007, indicated by the red vertical line.}
    \label{fig:gdp_forecast}
\end{figure}

\paragraph{Model.} We train a model on a GDP dataset with annual data, in which we produce 1-year ahead GDP growth forecasts based on 6 macroeconomic variables: growth in government expenditure, investment, consumption, exports, and imports, and real estate loans from commercial banks as a percent of GDP \cite{fred}. Our model first uses principal components analysis to reduce our data to 5 dimensions, then applies a K-nearest neighbors algorithm with 4 neighbors \cite{scikit-learn}. The number of principal components and neighbors were tuned using Scikit-learn's grid search, training on data from 1975-1990 and validating on data from 1991-2007. Our model achieves a validation $R^2$ of .26. Retraining the model on the full training dataset (1975-2007), we achieve a training $R^2$ of .40. However, our forecasts for 2008-2009 perform worse than chance: $R^2=-3.13$. From Figure \ref{fig:gdp_forecast}, it is clear that our model fails to forecast the financial crisis.

\paragraph{G-SHAP values.} We compute the G-SHAP values, where $g$ is a loss function. In our case, the loss function is $R^2$.

\begin{equation}
    g(f,X,\Omega) = L(f,X,y)
    = 1-\frac
    {\sum_t (\%\Delta\text{GDP}_t - f(x_{t-1}))^2}
    {\sum_t (\%\Delta\text{GDP}_t - \overline{\%\Delta\text{GDP}})^2}
\end{equation}

Where additional arguments to $g$ are the labels for $X$; $\Omega=\{y\}$.

\paragraph{Interpretation.} For interpretation, we make two comparisons. First, we compare the training data $X_\text{train}=\{x_{t-1} : t=1975...2007\}$ to a background dataset $Z$ (the shuffled training data) in terms of $g$ using training labels; $y_\text{train}=\{\%\Delta\text{GDP}_t : t=1975...2007\}$. Second, we compare the test data $X_\text{test}=\{x_{t-1} : t=2008,2009\}$ to $Z$ in terms of $g$ using test labels; $y_\text{test}=\{\%\Delta\text{GDP}_t : t=2008,2009\}$. By making these two comparisons, we can see the extent to which feature $j$ contributed to model performance for the training data, and the extent to which feature $j$ contributed to model performance for the test data. If feature $j$'s contribution diminished between the training and test data, we can conclude it was responsible for our model's poor test performance. A negative G-SHAP value indicates that feature $j$ degraded performance.

In the first comparison, our model's $R^2$ is .40 for the training data and -.28 for the background data (see Table \ref{tab:gdp_comparison}). Each G-SHAP value $\phi_g^j$ is the amount of this difference (.68) explained by feature $j$. We interpret the G-SHAP values for the test data comparison comparison similarly.

\begin{table}[]
    \centering
    \caption{Model performance for training (1975-2007) and test (2008-2009) data}
    \ra{1.3}
    \begin{tabular}{@{}lrr@{}}
        \toprule
        & \multicolumn{2}{c}{Years} \\
        \cmidrule{2-3}
        & 1975-2007 & 2008-2009 \\
        \midrule
        $g(f,X,\Omega)$ & 0.40 & -3.13 \\
        $g(f,Z,\Omega)$ & -0.28 & -3.94 \\
        $g(f,X,\Omega)-g(f,Z,\Omega)$ & 0.68 & 0.81 \\
        \bottomrule
    \end{tabular}
    \label{tab:gdp_comparison}
\end{table}

For model failure explanations more broadly, our model has a loss $L(f,X,y)$ for our test sample and $L(f,Z,y)$ for the background data. Each G-SHAP value $\phi_g^j$ is the amount of this difference $L(f,X,y)-L(f,Z,y)$ explained by feature $j$.

\begin{figure}
    \centering
    \includegraphics[scale=.7]{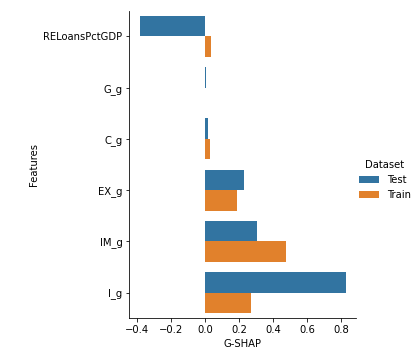}
    \caption{Normalized G-SHAP values for model performance on training data (1975-2007) and test data (2008-2009). The model's features 1-year lagged growth rates in government expenditure $G_g$, investment $I_g$, consumption $C_g$, exports $EX_g$, import $IM_g$, and 1-year lagged real estate loans as a percent of GDP $RELoansPctGDP$.}
    \label{fig:gdp_gshap}
\end{figure}

\paragraph{Results.} We plot our results in Figure \ref{fig:gdp_gshap}. This plot tells us which features contribute most to model performance. Because the performance of the background dataset differs in terms of $g$ when using the training versus test labels, we normalize the G-SHAP values to sum to 1 in order to compare performance. We see that real estate loans as a percent of GDP in 2007-2008 decreased the $R^2$ of our model's 2008-2009 GDP growth forecast by 38\%.

\newpage
\section{Broader impact}

It is increasingly important to explain how machine learning models work. Researchers have proposed a variety of methods to measure which features most affect a model's output. However, many important questions about a model cannot be answered just by measuring feature importance with respect to its output. Rather, we need to measure feature importance with respect to {\em some function} of the model's output. To answer this broader set of questions, we introduced Generalized Shapley Additive Explanations (G-SHAP) to compute generalized feature importances.

Explainability techniques such as G-SHAP may have negative impacts. For example, increased model transparency could exacerbate the threat of adversarial attacks. It is also possible that certain types of explanations, such as model failure explanations, could decrease user trust in reliable algorithms by drawing attention to rare errors.

G-SHAP also has significant potential for positive impacts. We illustrated these impacts by discussing three paradigmatic applications of G-SHAP: explaining {\em general classification} differences to improve transparency in medical diagnosis, explaining {\em intergroup differences} to detect potential sources of bias in models, and explaining {\em model failure} to predict a crisis in economic forecasting. Researchers and practitioners can easily extend our method to measure feature importance with respect to other generalized functions. We hope and expect that G-SHAP will provide a clean, principled way to measure feature importance for the purpose of improving algorithmic transparency, fairness, and performance.

\newpage
\bibliographystyle{unsrt}
\bibliography{references}

\begin{thebibliography}{10}

\bibitem{lipton2018mythos}
Zachary~C Lipton.
\newblock The mythos of model interpretability.
\newblock {\em Queue}, 16(3):31--57, 2018.

\bibitem{compas}
{\em Practitioners Guide to COMPAS}, 2012.

\bibitem{davenport2019potential}
Thomas Davenport and Ravi Kalakota.
\newblock The potential for artificial intelligence in healthcare.
\newblock {\em Future healthcare journal}, 6(2):94, 2019.

\bibitem{datta2015automated}
Amit Datta, Michael~Carl Tschantz, and Anupam Datta.
\newblock Automated experiments on ad privacy settings.
\newblock {\em Proceedings on privacy enhancing technologies}, 2015(1):92--112,
  2015.

\bibitem{deeplift}
Avanti Shrikumar, Peyton Greenside, and Anshul Kundaje.
\newblock Learning important features through propagating activation
  differences.
\newblock In {\em Proceedings of the 34th International Conference on Machine
  Learning-Volume 70}, pages 3145--3153. JMLR. org, 2017.

\bibitem{shap}
Scott~M Lundberg and Su-In Lee.
\newblock A unified approach to interpreting model predictions.
\newblock In {\em Advances in neural information processing systems}, pages
  4765--4774, 2017.

\bibitem{lime}
Marco~Tulio Ribeiro, Sameer Singh, and Carlos Guestrin.
\newblock "why should i trust you?" explaining the predictions of any
  classifier.
\newblock In {\em Proceedings of the 22nd ACM SIGKDD international conference
  on knowledge discovery and data mining}, pages 1135--1144, 2016.

\bibitem{shapley1953value}
Lloyd~S Shapley.
\newblock A value for n-person games.
\newblock {\em Contributions to the Theory of Games}, 2(28):307--317, 1953.

\bibitem{roth1988introduction}
Alvin~E Roth.
\newblock Introduction to the shapley value.
\newblock {\em The Shapley value}, pages 1--27, 1988.

\bibitem{cohen2005feature}
Shay~B Cohen, Eytan Ruppin, and Gideon Dror.
\newblock Feature selection based on the shapley value.
\newblock In {\em IJCAI}, volume~5, pages 665--670, 2005.

\bibitem{molnar2019}
Christoph Molnar.
\newblock {\em Interpretable Machine Learning}.
\newblock 2019.
\newblock \url{https://christophm.github.io/interpretable-ml-book/}.

\bibitem{lundberg2019explainable}
Scott~M Lundberg, Gabriel Erion, Hugh Chen, Alex DeGrave, Jordan~M Prutkin,
  Bala Nair, Ronit Katz, Jonathan Himmelfarb, Nisha Bansal, and Su-In Lee.
\newblock Explainable ai for trees: From local explanations to global
  understanding.
\newblock {\em arXiv preprint arXiv:1905.04610}, 2019.

\bibitem{ayush2020generating}
Kumar Ayush, Burak Uzkent, Marshall Burke, David Lobell, and Stefano Ermon.
\newblock Generating interpretable poverty maps using object detection in
  satellite images.
\newblock {\em arXiv preprint arXiv:2002.01612}, 2020.

\bibitem{singal2019shapley}
Raghav Singal, Omar Besbes, Antoine Desir, Vineet Goyal, and Garud Iyengar.
\newblock Shapley meets uniform: An axiomatic framework for attribution in
  online advertising.
\newblock In {\em The World Wide Web Conference}, pages 1713--1723, 2019.

\bibitem{jha2019attribution}
Susmit Jha, Sunny Raj, Steven Fernandes, Sumit~K Jha, Somesh Jha, Brian
  Jalaian, Gunjan Verma, and Ananthram Swami.
\newblock Attribution-based confidence metric for deep neural networks.
\newblock In {\em Advances in Neural Information Processing Systems}, pages
  11826--11837, 2019.

\bibitem{scikit-learn}
F.~Pedregosa, G.~Varoquaux, A.~Gramfort, V.~Michel, B.~Thirion, O.~Grisel,
  M.~Blondel, P.~Prettenhofer, R.~Weiss, V.~Dubourg, J.~Vanderplas, A.~Passos,
  D.~Cournapeau, M.~Brucher, M.~Perrot, and E.~Duchesnay.
\newblock Scikit-learn: Machine learning in {P}ython.
\newblock {\em Journal of Machine Learning Research}, 12:2825--2830, 2011.

\bibitem{propublica}
Jeff Larson, Surya Mattu, Lauren Kirchner, and Julia Angwin.
\newblock Machine bias.
\newblock 2016.

\bibitem{feldman2015certifying}
Michael Feldman, Sorelle~A Friedler, John Moeller, Carlos Scheidegger, and
  Suresh Venkatasubramanian.
\newblock Certifying and removing disparate impact.
\newblock In {\em proceedings of the 21th ACM SIGKDD international conference
  on knowledge discovery and data mining}, pages 259--268, 2015.

\bibitem{pan2010transfer}
S.~J. {Pan} and Q.~{Yang}.
\newblock A survey on transfer learning.
\newblock {\em IEEE Transactions on Knowledge and Data Engineering},
  22(10):1345--1359, 2010.

\bibitem{chen2018lifelong}
Zhiyuan Chen and Bing Liu.
\newblock Lifelong machine learning.
\newblock {\em Synthesis Lectures on Artificial Intelligence and Machine
  Learning}, 12(3):1--207, 2018.

\bibitem{fred}
U.S.~Bureau of~Economic~Analysis.
\newblock Real gross domestic product [gdpc1], real government consumption
  expenditures and gross investment [gcec1], real gross private domestic
  investment [gpdic1], real personal consumption expenditures [pcecc96], real
  exports of goods and services [expgsc1], real imports of goods and services
  [impgsc1], real estate loans, all commercial banks [relacbw027sbog], 2020.
\newblock data retrieved from Federal Reserve Bank of St. Louis,
  \url{https://fred.stlouisfed.org/}.

\end{thebibliography}

\end{document}